\documentclass[10pt, a4paper]{article}

\usepackage[final]{lrec2026} % this is the new style
\usepackage{newunicodechar}
\newunicodechar{Ə}{\scalebox{1.2}[1.2]{ə}}

\title{Keyboards for the Endangered Idu Mishmi Language}

\name{Akhilesh Kakolu Ramarao}
\address{Department of English Language and Linguistics \\ Faculty of Arts and Humanities, Heinrich Heine University Düsseldorf \\  akhilesh.kakolu.ramarao@uni-duesseldorf.de}

\abstract{
We present mobile and desktop keyboards for Idu Mishmi, an endangered Trans-Himalayan language spoken by approximately 11,000 people in Arunachal Pradesh, India. A Latin-based orthography, the Idu Azobra, was developed in 2018, but no digital input tools existed to use it. The orthography requires characters absent from standard keyboards, including schwa (ə), retracted vowels (ə̱, o̱, u̱), nasalized vowels, and accented forms, several of which involve multi-codepoint Unicode sequences that default keyboards do not support. Developed with the Idu Mishmi community, the keyboards comprise: (1) an Android mobile keyboard, published on the Google Play Store, and (2) a Windows desktop keyboard distributed as a single portable executable. Both tools support the complete character inventory, and operate fully offline with zero network permissions. The Android keyboard has been adopted by community leaders and teachers who currently know and actively use the Idu Azobra orthography. The Windows keyboard is currently undergoing testing with community leaders. We describe the design, implementation, and deployment as a replicable model for other endangered language communities.
\\ \newline }

\begin{document}

\maketitleabstract

\section{Introduction}

Idu Mishmi, also known as Kera'a \citep{peck2020keraa} (ISO 639-3: \texttt{clk}), is an endangered Trans-Himalayan language spoken in five districts of Arunachal Pradesh, India, mainly in the Dibang Valley and Lower Dibang Valley regions. The number of speakers is estimated at 11,000 \citep{lewis2015ethnologue}, covering two main dialects, Midu and Mithu \citep{reinohl2022locating}. The language is endangered for social, economic, and political reasons: Hindi and English serve as prestige varieties that speakers shift to \citep{kaland2023tone}, and Idu Mishmi is mostly no longer the primary language of the home. For the community, this shift threatens not only the language but the cultural identity embedded in it, a concern shared by many language communities across northeast India.

Community-led efforts to reverse this shift have produced significant results. The Idu Language Development Committee (ILDC) formally developed the Latin-based \textit{Idu Azobra}, the official and standard script for Idu Mishmi, in 2018. That same year, the first Idu Mishmi dictionary was completed \citep{blench2018dictionary}, comprising both dialects, alongside a grammar \citep{blench2018grammar} and the school textbook \textit{Asahi}. In 2021, Idu Mishmi entered classrooms as an optional third language in Idu-inhabited districts.

However, while the orthography exists on paper, no digital input tools existed to use it. The Idu Azobra requires characters such as schwa (ə), retracted vowels (ə̱, o̱, u̱), and their accented combinations, none of which are available on most default keyboards. Without a way to type their language, teachers cannot prepare instructional materials, community organizations cannot write official communications, and young speakers cannot use Idu Mishmi in the digital spaces where they increasingly communicate. 

We address this gap with two tools, developed in collaboration with the Idu Mishmi Cultural and Literary Society (IMCLS), the community's apex body for language and cultural affairs: (1) an \textbf{Android mobile keyboard}, published on the Google Play Store and actively used in teacher training programs, and (2) a \textbf{Windows desktop keyboard} currently undergoing community testing. The character inventory, layout decisions, and testing were carried out jointly with community leaders and teachers, ensuring that the tools reflect the orthography as sanctioned by the community. Both keyboards support the full Idu Mishmi character inventory, operate entirely offline, and are open-source. We describe the design, and deployment, and present this as a model that may be adapted by other endangered language communities.

\section{Background}
\label{sec:background}

\subsection{The Idu Mishmi Writing System}

The Idu Azobra uses Latin letters extended with special characters to represent sounds that English does not have \citep{blench2015writing}. These include the schwa (ə), retracted vowels (ə̱, o̱, u̱) where the tongue root is pulled back, five nasalized vowels (ã, ẽ, ĩ, õ, ũ), and three accent marks for tonal contrasts: grave for low tone, acute for high tone, and macron for mid tone. Table \ref{tab:characters} lists the full set of characters that need dedicated keyboard support.

\begin{table}[ht]
\centering
\small
\setlength{\tabcolsep}{4pt}
\begin{tabular}{@{}lll@{}}
\hline
\textbf{Category} & \textbf{Characters} & \textbf{Unicode sequence} \\
\hline
Schwa        & ə, Ə   & U+0259, U+018F \\
Retracted ə  & ə̱, Ə̱   & base + U+0331 \\
Retracted o  & o̱, O̱   & base + U+0331 \\
Retracted u  & u̱, U̱   & base + U+0331 \\
\hline
Grave        & à, è, ì, ò, ù   & precomposed (e.g. U+00E0) \\
Acute        & á, é, í, ó, ú   & precomposed (e.g. U+00E1) \\
Nasalized    & ã, ẽ, ĩ, õ, ũ   & precomposed (e.g. U+00E3) \\
Macron       & ā, ē, ī, ō, ū   & precomposed (e.g. U+0101) \\
\hline
Accented ə   & ə̀, ə́, ə̃, ə̄     & U+0259 + accent \\
Accented ə̱   & ə̱̀, ə̱́, ə̱̃, ə̱̄     & U+0259 + U+0331 + accent \\
Accented o̱   & ò̱, ó̱, õ̱, ō̱     & o + U+0331 + accent \\
Accented u̱   & ù̱, ú̱, ũ̱, ū̱     & u + U+0331 + accent \\
\hline
\end{tabular}
\caption{Idu Mishmi characters requiring keyboard support. U+0331 is Combining Macron Below (retraction); accents are U+0300 (grave), U+0301 (acute), U+0303 (tilde/nasalization), U+0304 (macron).}
\label{tab:characters}                                        
\end{table}

In Unicode, common accented vowels like à (U+00E0) or ẽ (U+1EBD) each have a single precomposed codepoint. However, accented schwas must be built by combining separate codepoints: for example, ə̀ is U+0259 (Latin Small Letter Schwa) followed by U+0300 (Combining Grave Accent). Retracted vowels use U+0331 (Combining Macron Below): for example, ə̱ is U+0259 + U+0331. Accented retracted schwas require three codepoints in canonical order: ə̱̀ is U+0259 + U+0331 (CCC,220) + U+0300 (CCC,230), where lower combining class values must come first. Getting this order wrong produces text that looks identical on screen but is treated as different by software, causing failures in search and copy-paste. Both keyboards enforce this ordering for all output.

\subsection{Related Work}

Keyboards are recognized as foundational infrastructure for digital language use, and community control over such technology is essential for meaningful revitalization \citep{meighan2021decolonizing}. Several projects have developed keyboards for endangered languages, including Livonian \citep{hamalainen2021livonian} and Plains Cree syllabics \citep{santos2020plains}. A recurring design challenge is what \citet{paterson2015keyboard} calls the ``diacritic density problem'': tone languages use accent marks far more often than European languages, and standard keyboards are not designed to handle this. Idu Mishmi shares this challenge, as a single vowel can carry a tone mark, nasalization, and retraction at the same time.

On the desktop side, existing tools for building custom keyboards each have drawbacks for our setting. Microsoft Keyboard Layout Creator (MSKLC) can only output one Unicode code per key, so it cannot produce characters like ə̱ that require two codes. Windows Text Services Framework (TSF) integrates with the system language bar but requires administrator privileges to install, which is not feasible on shared school and government computers. Keyman\footnote{\url{https://github.com/keymanapp/keyman}} is the most capable alternative: it supports multi-codepoint output, dead key composition, cross-platform deployment, and is widely used for endangered language keyboards. However, Keyman requires users to install a runtime engine, and on shared school and government computers where users lack installation privileges, this creates a barrier. Our hook-based approach offers a single executable that runs without requiring administrator access.  A longer-term path is to submit an Idu Mishmi keyboard layout to the Unicode CLDR \footnote{\url{https://cldr.unicode.org/}}(Common Locale Data Repository). Layouts accepted into CLDR can eventually be shipped natively by operating systems, removing the need for any third-party software. However, CLDR inclusion requires a stable, community-approved layout and a formal locale definition, both of which are preconditions we are working toward.

\section{System Design}
\label{sec:design}

\subsection{Android Mobile Keyboard}
\label{sec:android}

The Android keyboard is built on HeliBoard,\footnote{\url{https://github.com/Helium314/HeliBoard}} an open-source, privacy-focused keyboard app (GPL-3.0). Rather than adding Idu Mishmi support to HeliBoard directly, we released a separate app for three reasons: (1) a community-branded app is easier to find on the Play Store and builds trust with speakers; (2) Idu Mishmi-specific settings, such as which characters appear under each key, are hardcoded rather than left as user options; and (3) the community retains full control over updates without depending on the upstream project's release schedule. Building on an established keyboard gave us text prediction, gesture typing, clipboard management, and theming out of the box, so we could focus entirely on Idu Mishmi customizations.

\begin{figure*}[t]
\centering
\includegraphics[height=3.3cm]{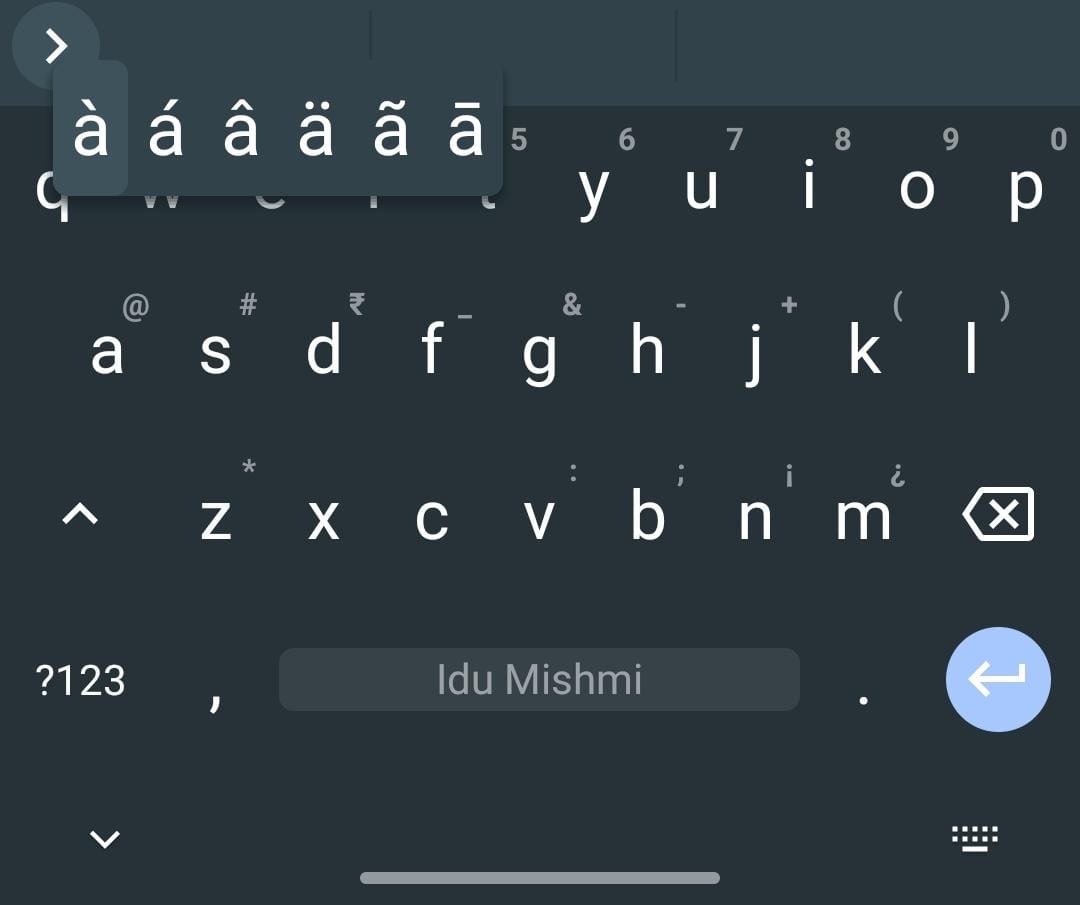}%
\hspace{1em}%
\includegraphics[height=3.3cm]{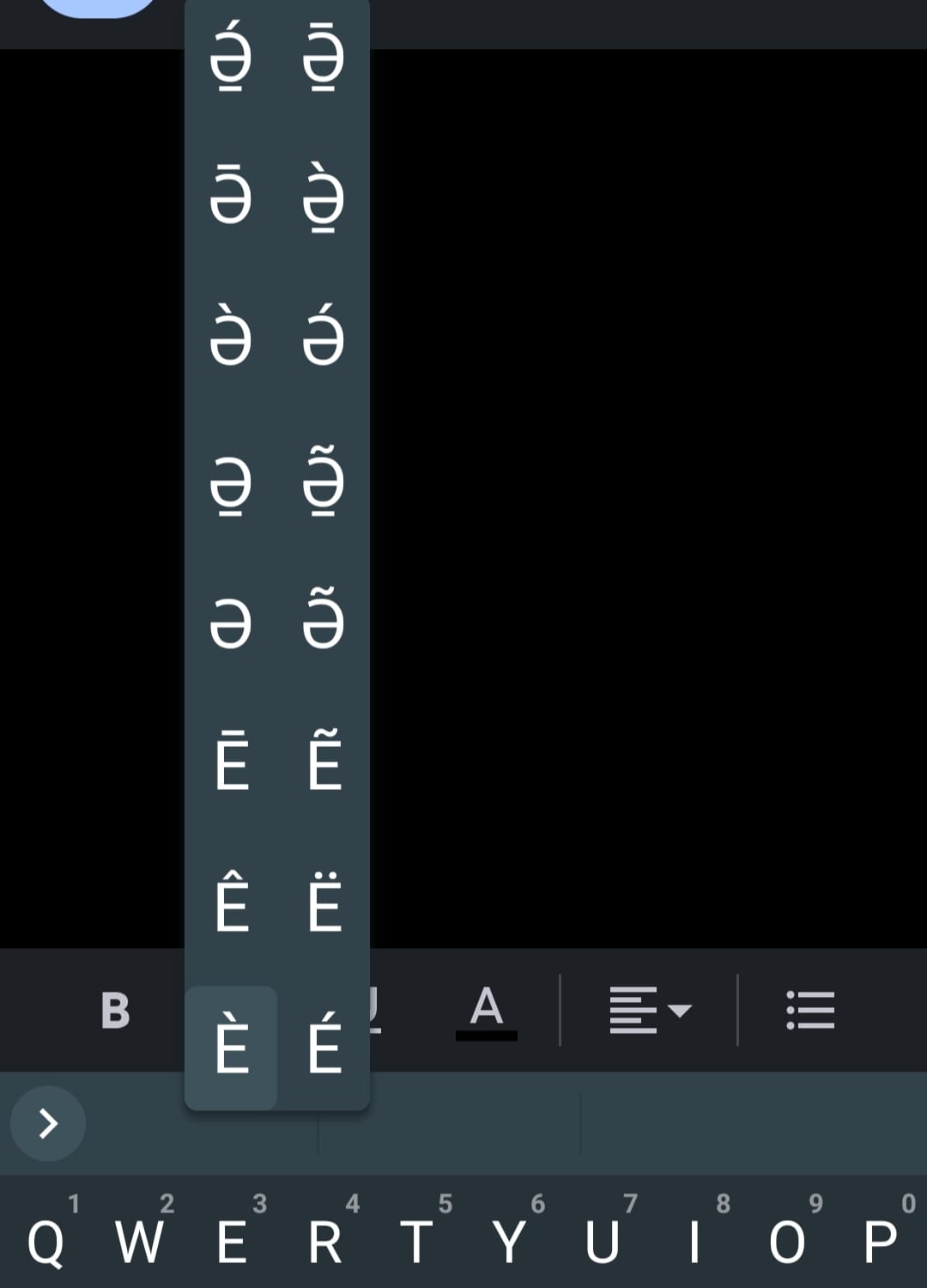}
\caption{The Idu Mishmi Android keyboard. (Left) Long-pressing \texttt{a} reveals accented vowel variants. (Right) Long-pressing \texttt{e} reveals the schwa inventory, including retracted schwa (ə̱) and accented forms, alongside standard accented \texttt{e} variants.}
\label{fig:keyboard}
\end{figure*}

\paragraph{Layout and input method.} The keyboard uses a standard QWERTY layout, since Idu Mishmi speakers are bilingual in English. To type Idu Mishmi characters, the user \textbf{long-presses} a base key, which opens a popup strip of related characters (Figure \ref{fig:keyboard}). For example, long-pressing \texttt{e} reveals all schwa variants (ə, ə̱) and their accented forms alongside standard accented \texttt{e} characters. Similarly, \texttt{o} and \texttt{u} include retracted o̱ and u̱ with their accented forms, while \texttt{a} and \texttt{i} provide accented and nasalized variants. This grouping was decided in consultation with community members, with the most frequently used Idu-specific characters placed at the beginning of each strip for quick access. No memorization is needed; users simply browse the popup visually, making the keyboard accessible to first-time users. We registered the app with a locale code for Idu Mishmi, so that Android recognizes it as an Idu Mishmi input method. The current release uses English dictionaries for basic word completion; an Idu Mishmi dictionary is planned as future work (see Section \ref{sec:conclusion}).

\paragraph{Privacy and compatibility.} The app requests no internet permissions; all text processing is on-device. It targets SDK 35 with minimum SDK 21, covering approximately 98\% of active Android devices.

\subsection{Windows Desktop Keyboard}
\label{sec:windows}

The Windows keyboard is a standalone program written in Go, distributed as a single portable executable that requires no additional software. It works by intercepting keystrokes at the operating system level using a low-level keyboard hook (\texttt{WH\_KEYBOARD\_LL}). When the user presses a mapped key combination, the program suppresses the original keystroke and sends the corresponding Idu Mishmi character to whichever application is active. Normal typing, including standard shortcuts (Ctrl+C, Alt+Tab, etc.), is unaffected.

\paragraph{Input method.} The keyboard provides Idu Mishmi characters through the \textbf{AltGr} key (Right Alt) in two ways. First, \textit{direct mappings} produce special characters in a single step: AltGr+E $\rightarrow$ ə, AltGr+R $\rightarrow$ ə̱, AltGr+O $\rightarrow$ o̱, AltGr+U $\rightarrow$ u̱ (with Shift for uppercase). Second, \textit{dead keys} allow two-step composition for accented characters: the user first presses AltGr plus an accent key (backtick for grave, apostrophe for acute, tilde for nasalization, hyphen for macron), then presses a vowel key to produce the accented form. For example, AltGr+\textasciitilde{} followed by AltGr+E produces ə̃. Dead keys compose with both plain and retracted schwas. The implementation ensures that multi-codepoint characters are always output in the correct Unicode order.

\paragraph{Compatibility with other input methods.} The keyboard coexists with other input methods (e.g., Hindi InScript, Devanagari phonetic) without conflict, as it only activates on AltGr combinations and passes all other keystrokes through unchanged. This is important in practice as many Idu Mishmi speakers also type in Hindi and English, and the keyboard must not interfere with those workflows.

\paragraph{Limitations.} The hook-based approach was chosen for portability but has known limitations. Some enterprise security policies may block low-level hooks. When deleting a composed character like ə̱̀ (which consists of three Unicode codepoints), most modern applications remove the entire character in one backspace press, but some older applications delete one codepoint at a time, requiring multiple presses. The keyboard guarantees correct ordering for new input but cannot fix incorrectly ordered sequences in pasted text, as it has no access to the text field itself.

\section{Deployment and Community Impact}
\label{sec:deployment}

The Android keyboard was launched in December 2025 and has approximately 100 installations. Its users are mainly community leaders and teachers, the people who currently know and actively use the Idu Azobra orthography. This reflects where orthography adoption stands today: since 2021, Idu Mishmi has been taught as an optional third language in Idu-inhabited districts, and the first textbook, \textit{Asahi (Idu Azobra Level I)}, was developed by the Idu Tho Ahie Committee (ITAC) under the IMCLS and submitted to the State Council of Educational Research and Training (SCERT). Community leaders now use the keyboard to train school teachers in digital literacy for this program, and the user base is expected to grow as more people learn the orthography. The Windows keyboard is being tested with community members, targeting schools and government offices where Windows desktops are standard.

Both tools work fully offline. This is necessary because internet connectivity in the Dibang Valley is limited, and it is also a deliberate choice to keep language data under community control rather than sending it to external servers \citep{meighan2021decolonizing}. For Idu Mishmi speakers living in other Indian cities, the Play Store listing makes the keyboard easy to find, and the offline design means it works the same way everywhere.

\paragraph{Sustainability and governance.} Long-term maintenance is managed by the IMCLS, which has decision-making authority over the orthography and its digital tools. The source code is publicly available under GPL-3.0, so the community is not dependent on any single developer. Teacher training sessions, run as part of the Idu Azobra school program, include instruction on installing and using the keyboard, making it part of existing educational infrastructure rather than a separate tool. Adopting the keyboards in schools and government offices requires no special IT setup: the Windows program is a single portable file that needs no installation or administrator access, and the Android app is on the Play Store.

\section{Conclusion and Future Work}
\label{sec:conclusion}

We have presented the first mobile and desktop keyboards for Idu Mishmi, bridging the gap between the 2018 orthography development and the digital tools needed to use it. Together, they support the complete Idu Azobra character inventory on the two platforms speakers use most: Android phones for daily communication and Windows desktops in schools and offices.

The approach is designed to be replicable by other endangered language communities with modest resources. It requires four steps: (1) identify the language's character inventory, working with the orthography committee; (2) for mobile, fork an open-source keyboard such as HeliBoard, modifying layout and locale configuration files; (3) for desktop, build a lightweight input method using OS-level keyboard hooks; and (4) engage the community in testing and deployment.

We identify four directions for future work.

\textit{Usability evaluation.} We need to measure how well the keyboard works in practice: typing speed (words per minute), how many key presses each character requires, and how often users make errors. Because the keyboard collects no usage data by design, this study must be done in person with participants in the Dibang Valley. The Android keyboard has been available for only a few months at the time of writing, and reaching a remote region takes time; we plan to carry out this evaluation as the user base grows.

\textit{Dictionary and text prediction.} The current release does not include an Idu Mishmi dictionary, so next-word prediction and autocorrect are not yet available. Adding one is not straightforward: Android's built-in dictionary engine strips accent marks when matching words (for example, it treats à and a as the same character). This works for most languages, but in Idu Mishmi the accents carry meaning, as grave and acute mark different tones, and nasalization is distinctive. Stripping them would merge words that are actually different. We will need to either modify the engine to preserve accents during matching, or use a dictionary format that keeps the full Unicode characters intact. The available dictionary \citep{blench2018dictionary}, with approximately 3,500 entries across the Midu and Mithu dialects, would also need decisions about how to handle compound words, dialectal variants, and tone-distinguished word pairs before conversion.

\textit{Font coverage and packaging.} Not all fonts installed on users' devices support the full Idu Mishmi character set, particularly the capital schwa (Ə, U+018F) and combining sequences. We plan to develop font coverage recommendations for common platforms and explore bundling a suitable open-source font (e.g., Noto Sans) with the keyboards to ensure consistent rendering.

\textit{Corpus building.} Finally, we plan to explore opt-in, privacy-respecting ways to build an Idu Mishmi text corpus from keyboard usage, which would support further language technology research.

\section*{Acknowledgements}

We thank the Idu Mishmi Cultural and Literary Society (IMCLS) for the opportunity to collaborate on this project. We are particularly grateful to Mite Lingi and Hindu Meme, without whom this work would not have been possible. We also thank Julia Muschalik for valuable brainstorming discussions, and Annu Kurian Mathew, Akash Ramaprasad and Apoorva Kakolu Ramarao for helping with testing.

\section*{Language Resource References}

\begin{itemize}
    \item Android keyboard codebase: \url{https://anonymous.4open.science/r/idu_keyboard-59C3/}
    \item Windows keyboard codebase: \url{https://anonymous.4open.science/r/idu-windows-keyboard-773E/}
\end{itemize}

\section{Bibliographical References}

\bibliographystyle{lrec2026-natbib}
\bibliography{references}

\end{document}